\documentclass[letterpaper, 10 pt, conference]{ieeeconf}  

\IEEEoverridecommandlockouts                              

\usepackage{graphicx} 
\usepackage[caption=false, font=footnotesize]{subfig}
\usepackage{booktabs}
\usepackage{amsmath} 
\usepackage{amssymb}  
\usepackage{algorithm}
\usepackage[noend]{algpseudocode}

\graphicspath{ {figures/} }

\DeclareMathOperator{\E}{\mathbb{E}}

\title{\LARGE \bf
Perception as prediction using general value functions in autonomous driving applications
}

\author{Daniel Graves$^{*}$,
    \thanks{$^*$Huawei Technologies Canada, Ltd., Edmonton, Canada {\tt\small daniel.graves@huawei.com}}
    Kasra Rezaee$^\dag$, and 
  	\thanks{$^\dag$Huawei Technologies Canada, Ltd., Markham, Canada {\tt\small kasra.rezaee@huawei.com}}
    Sean Scheideman$^\ddag$
    \thanks{$^\ddag$University of Alberta, Edmonton, Canada {\tt\small searn@ualberta.ca}}
}

\begin{document}
\maketitle
\thispagestyle{empty}
\pagestyle{empty}

\begin{abstract}

We propose and demonstrate a framework called perception as prediction for autonomous driving that uses general value functions (GVFs) to learn predictions.  Perception as prediction learns data-driven predictions relating to the impact of actions on the agent's perception of the world.
It also provides a data-driven approach to predict the impact of the anticipated behavior of other agents on the world without explicitly learning their policy or intentions.
We demonstrate perception as prediction by learning to predict an agent's front safety and rear safety with GVFs, which encapsulate anticipation of the behavior of the vehicle in front and in the rear, respectively.
The safety predictions are learned through random interactions in a simulated environment containing other agents.
We show that these predictions can be used to produce similar control behavior to an LQR-based controller in an adaptive cruise control problem as well as provide advanced warning when the vehicle behind is approaching dangerously.
The predictions are compact policy-based predictions that support prediction of the long term impact on safety when following a given policy.
We analyze two controllers that use the learned predictions in a racing simulator to understand the value of the predictions and demonstrate their use in the real-world on a Clearpath Jackal robot and an autonomous vehicle platform.

\end{abstract}

\section{Introduction}
Understanding the world by learning predictions and using those predictions to act intelligently in the world is becoming an important topic of research, cf \cite{horde2011}\cite{pilarski2011}\cite{gunther2016}\cite{kahn2018}\cite{dosovitskiy2016}\cite{modayil2012}.
Modern theory of the brain shows that we are predictive machines that constantly try to match incoming sensory inputs with predictions \cite{clark2013}.
An important highlight of this work are that actions and sensory perception are intimately related suggesting that when building intelligent predictive machines, learning to make action-oriented predictions could be a valuable way for an machine to understand and interact with the world.
The Horde framework embraces this idea of predicting sensorimotor signals \cite{horde2011} using general value functions to learn and make action and policy dependent predictions where a robot is able to learn to predict quantities like time-to-stop and time-to-obstacle by exploring and interacting with its environment.  This idea was extended to learn predictions at multiple time scales \cite{modayil2012}.
There have been a number of successful applications of general value functions (GVFs) including controlling a myoelectric robotic arm prosthesis \cite{pilarski2011} and controlling a laser welding robot from a camera \cite{gunther2016}.

Recently, action-conditioned predictions were learned and used to navigate a robot with a camera \cite{kahn2018} improving upon a similar technique introduced in \cite{dosovitskiy2016}.
They proposed supervised learning to learn to predict future event cues in the environment off-policy, such as through random exploration in the environment.
They used existing detection methods to label the event cues.
Our work distinguishes from composable action-conditioned predictors \cite{kahn2018} in several ways.
Firstly, we learn to make predictions with temporal difference (TD) learning \cite{sutton1988} instead of supervised learning.
The advantage of learning policy-based predictions with TD is that they are easier to learn while a composable action-conditioned predictor must generalize to all possible action sequences and policies.
In their work, an action sequence is needed to make long term predictions; however, in our work, the predictions are policy-based predictions that predict the accumulated value after executing the policy to completion.
In \cite{djuric2018}, trajectories of other traffic actors are predicted using deep learning from an image.
While a powerful approach, it can be difficult to learn in comparison to predicting safety with GVFs; we argue that predicting the trajectory of the traffic actors is not required.
Instead, we advocate predicting the impact of an action taken by our agent on our safety without explicitly detecting and predicting the other traffic actors and their trajectories.
That is the prediction of the other traffic actors is implicit by directly connecting the state of the environment and actions of our (ego) agent directly with the goal of evaluating and predicting our safety in a dynamic environment containing many traffic actors.

Control with GVFs though does not permit evaluating the safety of all possible policies; however, we hypothesize that making predictions about a small collection of policies can be a powerful way to control the vehicle.
The reason is that we ignore action trajectories that are unrealistic or impossible to execute.
As an example, a driver will not change the throttle, brake or steering very rapidly and thus most policies of interest are smooth; these policies embody the skills, or policies, that are available to our agent.
We demonstrate in our experiments that there is one target policy that is very useful in the adaptive cruise control problem: "what will happen if I keep doing what I'm doing?"
The intuition behind this special target policy is that it provides a signal that a controller can use to adjust its action.
We believe there are other useful policy-based predictions but we focus on learning and analyzing predictions with this target policy in this paper.

Ultimately, our goal is to bring action-oriented predictions to the autonomous vehicle that are expressive enough to be applied across many kinds of tasks including but not limited to (a) adaptive cruise control, (b) lane keeping, (c) lane change, (d) obstacle avoidance, (e) intersection navigation, and (f) parking.
In this work, we focus on learning predictions that will be analyzed and demonstrated on the task of adaptive cruise control (ACC).
The goal of ACC is to reduce rear-end collisions which account for nearly half of all vehicle crashes in the United States \cite{ntsb2015}\cite{acc_review2010}.
A number of ACC methods use predictive models, cf. \cite{corona2008}\cite{li2011}\cite{naus2010}\cite{magdici2017}, however the predictive models employed are next state predictions which impose significant computational burden since the model has to be iterated many times to search for action sequences that minimize a cost function.
We desire long term predictions using temporal difference (TD) learning \cite{horde2011}\cite{modayil2012}\cite{sutton1988}\cite{adam2015} that are computationally efficient to learn and make while being a useful signal for control.

One important question that arises is what is the best way to use these predictions?
We highlight two possible controllers although we don't make claims that they are better than existing controllers.
Our desire is to demonstrate a new way to think about building autonomous vehicles that we hope is closer to the way we think humans drive a vehicle.
In addition, we hope that this work highlights some early building blocks that could be combined with other predictions and traditional perception outputs to build a scalable predictive control platform for autonomous driving that helps abstract some of the complexities of the environment when building the controllers.

One of the important contributions of this work is showing that we can make GVF predictions in the presence of other agents.
Previous works in general value functions \cite{horde2011}, \cite{pilarski2011}, \cite{gunther2016}, and \cite{modayil2012} do not focus on multi agent prediction problems like autonomous driving.
Because these predictions are learned from interactions in an environment with other agents, these predictions are potentially capable of understanding and implicitly predicting how other agents will act in the environment, and also react to our agent's actions.
For example, the safety predictions that we learn in this paper can learn to pick up cues in the behavior of the other agents in order to anticipate whether they will brake, and thereby reduce safety, or accelerate, and thereby increase safety.
Embedding this expectation of how another agent will act into the prediction of safety could be important for adapting an autonomous vehicle's behavior to different kinds of situations and drivers.

The contributions of this work are as follows:
\begin{itemize}
\item Advocating and demonstrating that learning and using action-conditioned policy-based predictions with GVFs in autonomous driving could be useful in better understanding the world and how an agent can impacted it by its actions
\item Learning predictions in a multi-agent environment that includes an expectation of the behavior of other vehicles in the environment and their reaction to our vehicle's behavior
\item Demonstrating action-conditioned policy-based predictions in adaptive cruise control and rear-end safety predictions
\end{itemize}

The rest of the paper is organized as follows:  section II describes the proposed architecture, section III introduces the training algorithms used to learn the GVFs, and section IV details our experiments.


\section{Perception as Prediction in Autonomous Driving}
\begin{figure}[thpb]
\centering
\includegraphics[width=5cm]{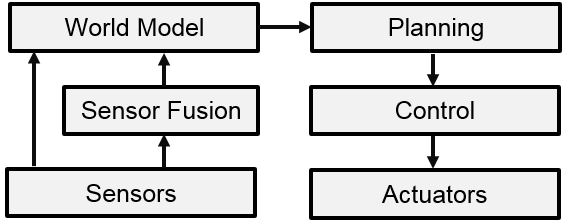}
\caption{Classical autonomous driving architecture}
\label{fig_classic_architecture}
\end{figure}

A common approach to designing an autonomous vehicle is to build layers to abstract the decision making that happens in planning and controlling a vehicle, cf. \cite{serban2018}\cite{zong2018}.
At a high level shown in figure \ref{fig_classic_architecture}, the planning layers use the world model generated by perception tasks to decide on a route, choose behaviors such as when to change lanes and then plan a path in the world.
This is then passed to a control layer that executes the plan produced by the planning layers including lateral (steering) and longitudinal control (throttle and brake).

\begin{figure}[thpb]
\centering
\includegraphics[width=6cm]{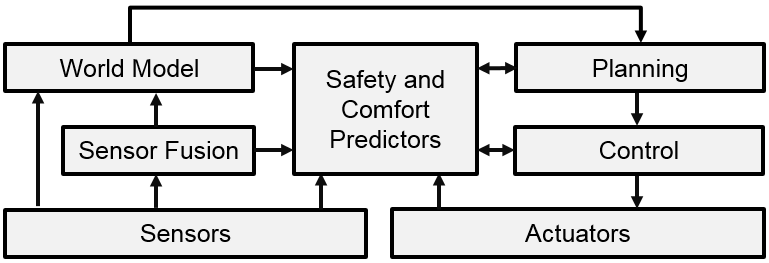}
\caption{Perception as prediction autonomous driving architecture}
\label{fig_predictive_architecture}
\end{figure}

We propose augmenting this architecture with action-oriented predictions as shown in figure \ref{fig_predictive_architecture}.

The key difference is that this architecture makes predictions with both sensors and actuator signals similar to modern understanding of the brain discussed in \cite{clark2013}.
This is an important distinction because it enables the autonomous agent to understand how the actions taken affect the sensor readings as well as the objects detected by a traditional perception pipeline.
This information is very helpful in allowing an agent to understand what actions to take in order to achieve a variety of goals in the environment.
What's beautiful about this approach is that when we learn a prediction, the prediction can be independent of the task in the environment and therefore represents part of the agent's knowledge of the world.
Therefore this architecture enables the application of action-oriented predictions that can potentially be applied to multiple tasks in the environment such as slowing down to avoid a collision with a slower vehicle or engaging obstacle avoidance behavior by either conducting an in-lane or out-of-lane maneuver.

We propose using GVFs first introduced in \cite{horde2011} to realize perception as prediction in autonomous driving.
We learn to predict our own speed, safety and safety of the vehicle behind us using GVFs and build two controllers using the predictions to understand the utility of these predictions:  a fuzzy controller and a rule-based controller.
We don't use the safety of the vehicle behind us for control purposes since it is unclear how to exploit that information for control purposes but we show that it can, for example, be used as an advanced warning system.
The predictions learned and demonstrated here can be combined with other predictions to achieve control solutions to more complex problems in autonomous driving.

\section{Methods}
\subsection{Safety Cumulants}
We start by defining safety; this is one of the important values we wish to predict with the GVF framework.
We borrow from a classical definition of safety in autonomous driving and adaptive cruise control using the inter-vehicle distance model also called headway, cf. \cite{magdici2017}\cite{moon2009}.
However, we extend this model to three dimensions to allow for use with high dimensional LIDAR sensors as well as low dimensional radar sensors.
The safety function is the pseudo-reward signal (also called the cumulant in general value function literature) which the predictor must learn to predict.
The safety cumulant function $c_f$ maps the current state $s_t$ at time $t$ to a ego safety score on the interval $[0,1]$.
There are two safety zones in our implementation:  ego safety (or front safety) and rear safety.
Front safety is a binary value defined by

\begin{equation}
  c_f(s_t) = \begin{cases}
    0, & \text{if }n_f > \beta_f.\\
    1, & \text{otherwise}.
  \end{cases}
  \label{eq_front_safety}
\end{equation}

where $n_f$ is the number of points returned by either LIDAR or radar sensors that are inside the front safety zone (or box) and $\beta_f \geq 0$ is a minimum threshold.
When using sensor fusion data to predict safety, $c_f(s_t)=0$ if a vehicle is detected in the front safety zone by the fusion module and is otherwise $c_f(s_t)=1$.
The width of the front safety zone is the width of the vehicle plus a safety margin and the height of the front safety zone is the height of the vehicle plus a safety margin for clearance.
The length of the front safety zone is the headway to the vehicle in front and is proportional to the vehicle's speed following the inter-vehicle distance model

\begin{equation}
h_f = d_\text{min} + v \tau
\label{eq_inter_vehicle_distance}
\end{equation}

where $\tau$ is the desired headway in seconds, $d_\text{min}$ is the stand-still distance, and $v$ is the current speed of the vehicle.
For rear safety, we build a three dimensional safety zone for the rear vehicle and calculate $h_r=d_\text{min} + v_r \tau$ where $v_r$ is the speed of the rear vehicle.

\subsection{Learning to Predict Safety with GVFs}
General value functions \cite{horde2011}\cite{modayil2012} form the basis of our perception as prediction framework to learn action-conditioned predictions of future safety.
Unlike \cite{horde2011} and \cite{modayil2012}, the predictions learned in this present work are action-conditioned predictions $Q^\pi(s,a)$ which are a function of state $s$ and action $a$ conditioned on policy $\pi$; we use TD($\lambda=0$) instead of GQ($\lambda$) to learn them.
Thus, the predictive question we aim to answer is "will I be safe if I take action $a$ and take similar actions thereafter?"
This predictive question is a predictive model of the world that permits queries over a set of possible next actions where subsequent actions are assumed to be similar.

The goal is to learn an estimator that predicts the return of the cumulant $G_t$ defined by 

\begin{equation}
G_t \equiv \sum_{k=0}^{\infty} {(\prod_{j=0}^{k} {\gamma_{t+j+1}}) c_{t+k+1}}
\label{eq_return}
\end{equation}

where $0 \leq \gamma_t<1$ is the continuation function and $c_{t}$ is the cumulant (pseudo-reward) sampled at time $t$.  The general value function is defined as 

\begin{equation}
Q^\pi(s,a) = \E_{\pi}[G_t | s_t=s,a_t=a,a_{t+1:T-1} \sim \pi, T \sim \gamma]
\label{eq_action_value}
\end{equation}

where $\pi$, $\gamma$, and $c$ make up the predictive question \cite{horde2011}.
Each GVF $Q^\pi$ is approximated with an artificial neural network parameterized by $\theta$ denoted as $\hat{q}^\pi(s,a,\theta)$.

Using non-linear function approximation introduces potential challenges in learning because there is no proof of convergence.
In addition, off-policy learning where the target policy $\pi$ may be different from the behavior policy $\mu$ could be problematic with deep neural networks if importance sampling ratios are required.

The approach adopted here uses TD($\lambda=0$) \cite{sutton1998} to learn the predictions using non-linear function approximation with an experience replay buffer.
The loss for the general value function $\hat{q}^\pi(s,a,\theta)$ is given by 

\begin{equation}
L(\theta) = \E_{s \sim d_\mu, a \sim \mu}[(y - \hat{q}^\pi(s,a,\theta))^2]
\label{eq_td_loss}
\end{equation}
where the target $y$ is produced by bootstrapping a prediction of the value of the next state and action taken from the target policy $\pi$ given by 

\begin{equation}
y=\E_{s' \sim P, a' \sim \pi}[c + \gamma \hat{q}^\pi(s',a',\theta)]
\label{eq_td_bootstrap}
\end{equation}

where $y$ is a bootstrap prediction using the most recent parameters $\theta$ but is ignored in the gradient descent.
$d_\mu$ is the state distribution of the behavior policy $\mu$ and $P$ is the Markovian transition distribution over next state $s'$ given state $s$ and action $a$.
The time subscript on $c$ and $\gamma$ has been dropped to simplify notation.
If the agent behaves with the same policy as the target policy $\pi$ such that $\mu(a|s)=\pi(a|s)$ then the approach is on-policy learning otherwise when the policies are different, it is off-policy.
Note that this approach doesn't correct for the state distribution $d_\mu$ because it still depends on the behavior policy $\mu$.
Both on-policy and off-policy approaches were tried but the behavior policy was constructed such that it was very similar to the target policy and thus no appreciable difference was noticed when learning on-policy or off-policy.

The gradient of the loss function \eqref{eq_td_loss} is given by
\begin{equation}
\nabla_{\theta} L(\theta) = \E_{s \sim d_\mu, a \sim \mu, s' \sim P}[\delta \nabla_{\theta} \hat{q}^\pi(s,a;\theta)]
\label{eq_td_gradient}
\end{equation}

where $\delta=y - \hat{q}^\pi(s,a;\theta)$ is the TD error.  When the agent observes a state $s$ and chooses an action $a$ according to its behavior policy, it receives cumulant $c$, continuation $\gamma$ and observes next state $s'$ and store this tuple in the replay buffer.
If the behavior policy is $\pi$, we also store the next action $a'$ that the agent takes in the replay buffer otherwise we generate an action $a' \sim \pi$ and store it in the replay buffer.
The experience stored in the replay buffer is given by the tuple $(s,a,c,\gamma, s',a')$.
When training the GVF, a mini-batch of $m<n$ samples, where $n$ is the size of the replay buffer, is sampled randomly to update the parameters $\theta$.
The updates can be either accomplished on-line, while collecting and storing experience in the replay buffer, or off-line.

\begin{algorithm}
\caption{GVF training algorithm}
\label{alg_gvf_train}
\begin{algorithmic}[1]
\State Initialize replay memory $D$ to capacity $n$
\State Initialize action-value function $\hat{q}^\pi$
\State Observe initial state $s_0$
\State Sample first action $a_0 \sim \mu$
\For{t=0,T}
  \State Execute action $a_t$ and observe state $s_{t+1}$
  \State Compute cumulant $c_{t+1}=c(s_t,a_t,s_{t+1})$
  \State Compute continuation $\gamma_{t+1}=\gamma(s_t,a_t,s_{t+1})$
  \State Sample target action $a' \sim \pi$
  \State Store transition $(s_t,a_t,c_{t+1},\gamma_{t+1},s_{t+1}, a')$ in $D$
  \State Sample random minibatch of transitions $(s_i,a_i,c_{i+1},\gamma_{i+1},s_{i+1}, a')$ from $D$
  \State Compute $y_i = c_{i+1} + \gamma_{i+1} \hat{q}^\pi(s_{i+1},a';\theta)$
  \State Perform gradient descent step on $(y_i-\hat{q}^\pi(s_i,a_i;\theta))^2$ according to \eqref{eq_td_gradient}
  \State Sample next action $a_{t+1} \sim \mu$
\EndFor
\end{algorithmic}
\end{algorithm}

The training algorithm is described in detail in algorithm \ref{alg_gvf_train}.
The algorithm is general for predicting the return of any cumulant; in this work, we learn separate estimators: one for each cumulant of future front safety, rear safety, and speed.
Predicting speed is accomplished with the current speed of the ego vehicle as the cumulant; the prediction is an average of future speeds of the vehicle across a defined temporal horizon.
Our motivation for predicting speed was that it was found to be slightly non-linear with different gear ratios and the fuzzy controller architecture in particular benefited from having predictions of speed when choosing an action to maintain safety.
The cumulants are scaled by a factor of $1-\gamma_{t+1}$ to normalize them; this ensures the predicted safety is on the interval $[0,1]$ since the sum of an infinite geometric series of $0 \leq \gamma<1$ is $1/(1-\gamma)$ if $\gamma$ is constant.

\subsection{Predictive Control}
In this section, we describe two very different controllers that are able to use the predictions of safety and speed to control the vehicle.
Both a fuzzy controller design and a rule-based controller design are introduced.
The objective in analyzing several controller designs in the simulator environment is to understand the utility of the predictions learned and ensure that the performance achieved was not in part due to a particular controller design.
The action space is different between each platform that we tested on and so there are small differences between the controllers in TORCS, on the Clearpath Jackal robot and on the autonomous driving platform.
These details will be discussed in the experimental section.

\subsubsection{Fuzzy Predictive Controller}
\begin{figure}[thpb]
\centering
\includegraphics[width=8.5cm]{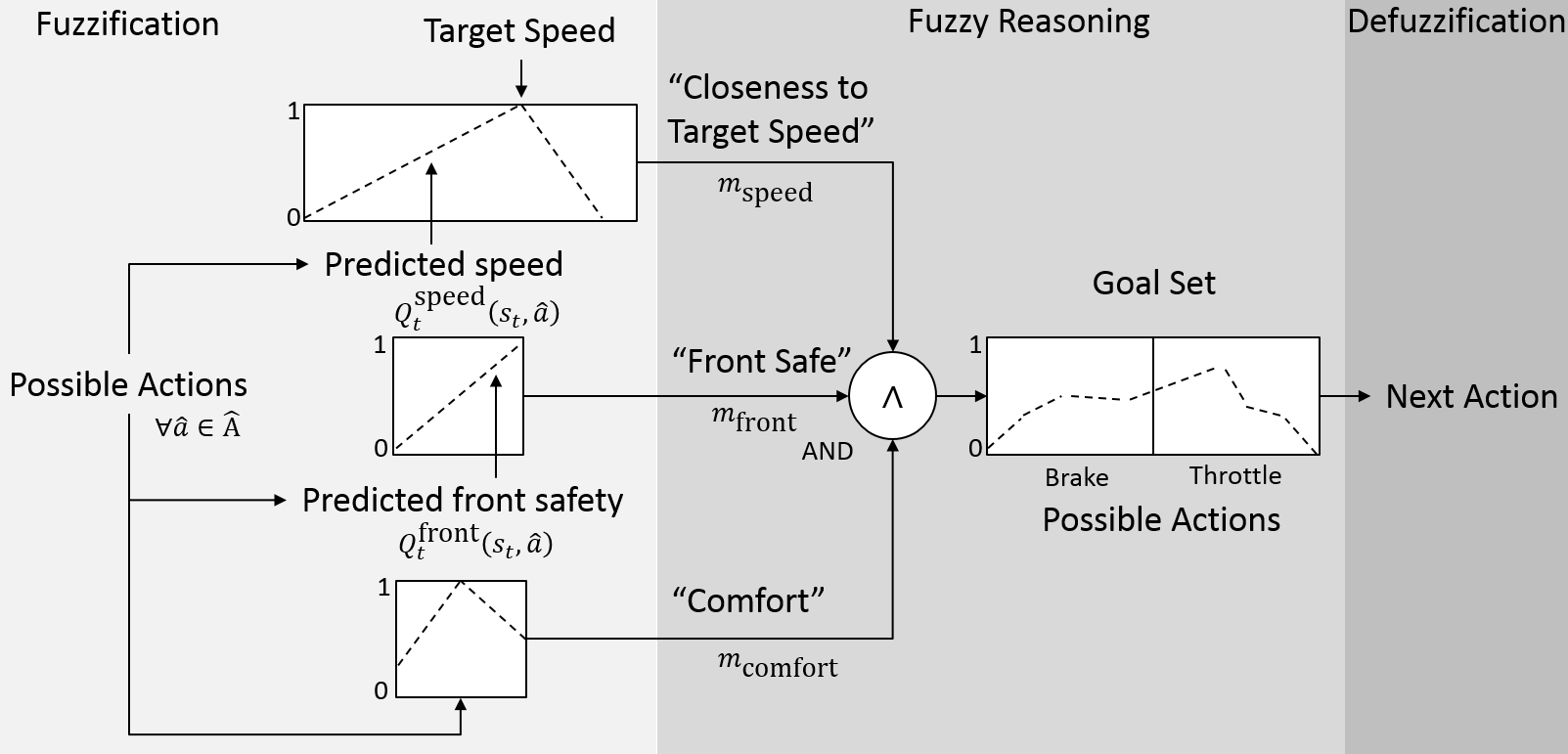}
\caption{Fuzzy controller architecture}
\label{fig_fuzzy_control}
\end{figure}

A classical fuzzy reasoning architecture was developed to test the predictions learned in TORCS:  a popular racing simulator.
The controller selects an action $\hat{a}_t$ given a collection of predictions about front safety and speed made by the GVFs.
This architecture requires some tuning of the fuzzy sets to achieve the desired behavior.
The controller aims to select actions that are front safe, close to target speed and comfortable.
The architecture is depicted in figure \ref{fig_fuzzy_control}.

There are fuzzy sets defined for each objective:  front safe, close to target speed and comfortable as depicted in figure \ref{fig_fuzzy_control}.
These are tuned manually to achieve the desired behavior.
The controller uses the predictions as models of safety and speed by sweeping across a set of possible next actions $\hat{a}_t \in \hat{A}$ given current state $s_t$ and producing predictions for each hypothetical action the agent could take.
These set of predictions are converted into a membership value through the fuzzification step that represents how closely the prediction achieves the desired safety, target speed, and comfort levels, respectively.
Comfort is defined as a fuzzy set over the action space emphasizing smaller throttle and brake actions as being more comfortable.
These fuzzy sets are combined into a single goal fuzzy set using the product t-norm operator.
The action was determined by calculating the centroid of the goal fuzzy set $g(a)$

\begin{equation}
a_t = \frac{\sum_{\forall \hat{a}_t \in \hat{A}} {g(\hat{a}_t)^m \hat{a}_t}}{\sum_{\forall \hat{a}_t \in \hat{A}} {g(\hat{a}_t)^m}}
\label{eq_fuzzy_centroid}
\end{equation}
where $m$ is a parameter that controls the greediness of the action selected.
Increasing this parameter improves the responsiveness of the controller.

\subsubsection{Rule-based Predictive Controller}
The second predictive controller is a rule-based design.
Two different rule-based designs were used depending on the environment with the primary difference being how the action space is defined.
For the first rule-based controller, the algorithm is given below.

\begin{algorithm}
\caption{Rule-based Safety Controller with Speed}
\label{alg_rule_torcs}
\begin{algorithmic}[1]
\Procedure{act($s_t$, $a_{t-1}$)}{}
\State Predict future safety $\hat{v}_\text{front}=\hat{q}^\pi_
\text{front}(s_t, a_{t-1})$
\State Predict future speed $\hat{v}_\text{speed}=\hat{q}^\pi_\text{speed}(s_t, a_{t-1})$
\If {$\hat{v}_\text{front} < \beta$}
  \State $a_t=a_{t-1}-\alpha_\text{decel}(1-\hat{v}_\text{front})$
\Else
  \State $e_\text{speed}=\text{max}(e_{min}, \text{min}(e_{max}, v_\text{target} - \hat{v}_\text{speed}))$
  \State $a_t=a_{t-1}+\alpha_\text{speed} e_\text{speed}$
\EndIf
\State \Return $\text{min}(a_\text{max}, \text{max}(a_\text{min}, a_t))$
\EndProcedure
\end{algorithmic}
\end{algorithm}

The parameter $\beta$ is the threshold that dictates when the controller starts to respond to unsafe situations.
The $\alpha_\text{decel}$ controls the rate of deceleration when the controller is unsafe.
Otherwise, the controller is a simple proportional controller that aims to achieve a target speed $v_\text{target}$.
The action space for this controller is typically throttle and brake.

A second rule-based controller was used when integrating the controller in both the robot and autonomous driving platform where there exists a control layer between adaptive cruise control module and the actuators.
In this case, the adaptive cruise control module must supply a target speed to the control layer to execute the command.

\begin{algorithm}
\caption{Rule-based Safety Controller without Speed}
\label{alg_rule_jackal}
\begin{algorithmic}[1]
\Procedure{act($s_t$, $a_{t-1}$)}{}
\State Predict future safety $\hat{v}_\text{front}=\hat{q}^\pi_
\text{front}(s_t, a_{t-1})$
\If {$\hat{v}_\text{front} < \beta_1$}
  \State $a_t=a_{t-1}-\alpha_\text{decel}(1-\hat{v}_\text{front})$
\ElsIf {$\hat{v}_\text{front} > \beta_2$}
  \State $a_t=a_{t-1}+\alpha_\text{accel}\hat{v}_\text{front}$
\Else
  \State $a_t=a_{t-1}$
\EndIf
\State \Return $min(max(a_t, 0.0), v_\text{target})$
\EndProcedure
\end{algorithmic}
\end{algorithm}

There are two safety threshold parameters for hysteresis such that $\beta_1 <= \beta_2$; whenever the prediction is between these two parameters, the action doesn't change.

\section{Experiments}
An analysis of the predictions and the control behavior of the controllers that use the predictions are provided in the TORCS environment under a number of different scenarios.
After evaluation, the rule-based controller was selected for implementation on the Jackal robot and the autonomous vehicle platform since it performed similarly with the fuzzy controller and was simpler to tune. 
A two stage approach is used: (1) learn the predictors by following the predictor's target policy and (2) use the predictors in autonomous driving applications (such as a warning system or adaptive cruise control).

\subsection{Training the Predictors}
The target policy chosen was the normal distribution centered on the last action taken, i.e. $\pi(a_t|s_t, a_{t-1})=\pi(a_t|a_{t_1})=\mathcal{N}(a_{t-1}, \sigma^{2})$ where $\sigma$ is a tunable parameter.
This target policy represents the question "what if I keep doing what I'm doing?"
The behavior policy chosen was a Wiener process where the next action is the last action plus noise generated by a normal distribution centered on the last action taken and with standard deviation equal to $\sigma$ for our target policy.
However, in order to facilitate exploration of the state and action space, the agent occasionally interrupts the Wiener process and chooses a random action according to uniform probability and then continues with the random walk. 
A large value of $\sigma$ is desirable to improve exploration of the state and action space but when choosing large $\sigma$, it can be challenging to learn longer term predictions since the actions are changing too rapidly.

We trained predictions for a number of different $\gamma$ values including 0.95, 0.975 and 0.983 which correspond to approximately 1 second, 2 second and 3 second prediction horizons.
When training the safety predictors, other vehicles are needed on the road in order to predict their impact on our safety.
Therefore, the training terminates with $\gamma=0$ upon collision with another vehicle.
When training the speed predictor, we do not train with other vehicles on the road and so there is no termination condition.

\subsection{Analysis of TORCS Experiments}
We experimented on a number of different scenarios but here we highlight the two most challenging high-speed scenarios: (a) emergency stop and (b) follow-and-stop.
The target speed in both of these experiments was 100 km/h.
In the first scenario the vehicle approaches a stopped vehicle and must stop quickly to avoid collision.
In the second scenario the vehicle follows a slower vehicle going 80 km/h which then abruptly stops requiring the vehicle to react and slow down without a collision.
The LQR baseline solution described in \cite{moon2009}, called ACC/CA, was implemented in TORCs as a representative for the state of the industry.
ACC/CA is a full-range adaptive cruise control system that aims to completely avoid rear-end collisions when vehicle-following in unsafe situations while driving comfortably during normal driving conditions.
The system parameters are tuned by using real-driving data.
We chose to compare to ACC/CA because of the additional driving control strategies for avoiding collisions in unsafe situations when safety becomes more important than comfort.
The target safety parameters for all controllers were defined by a desired spacing of $\tau=3$ seconds, minimum stopping distance of $d_\text{min}=4$ meters and $\beta_f=0$.
The parameters of the controllers were tuned in order to achieve similar or better performance than the baseline.
The objective was not to beat the baseline but to match its performance and show that the predictions can be used to control the vehicle.

The sensor data used to predict safety was the distance to the vehicle front, the change in the distance, the previous change in the distance, the speed of the ego vehicle, and percentage of throttle and brake.
The speed predictor was only learned and used in TORCS and we used the current speed, the engine RPM, gear ratio and percentage of throttle and brake as inputs.

In the first set of experiments, the performance of the predictions is analyzed by comparing them with the safety signals being predicted.
The ACC/CA baseline controller is used to drive the vehicle.

\begin{figure*}[t]
	\centering
    \subfloat[Front safety predictions]
    {
  	\includegraphics[width=5.5cm]{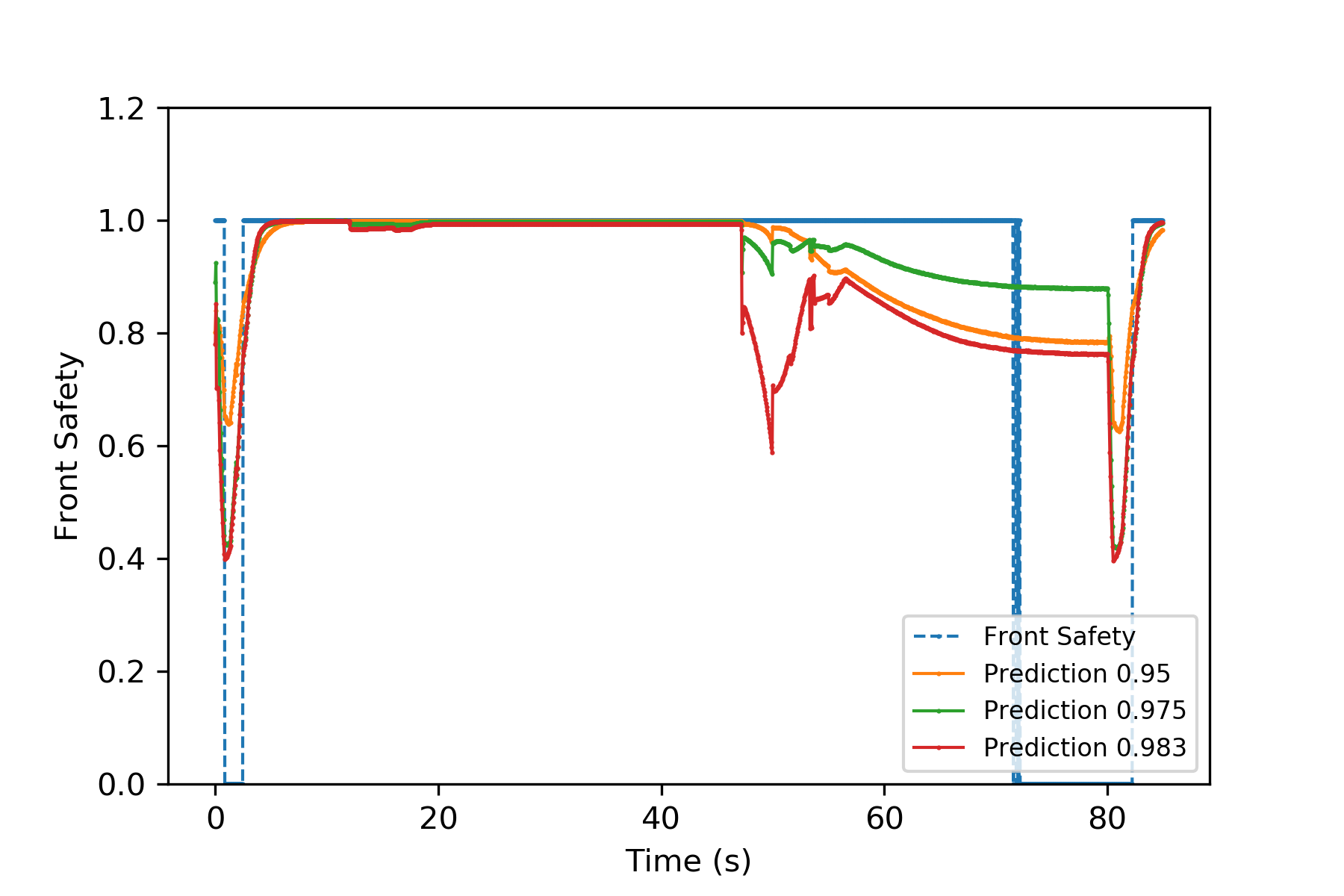}
    }
	\subfloat[Back safety predictions]
	{
	\includegraphics[width=5.5cm]{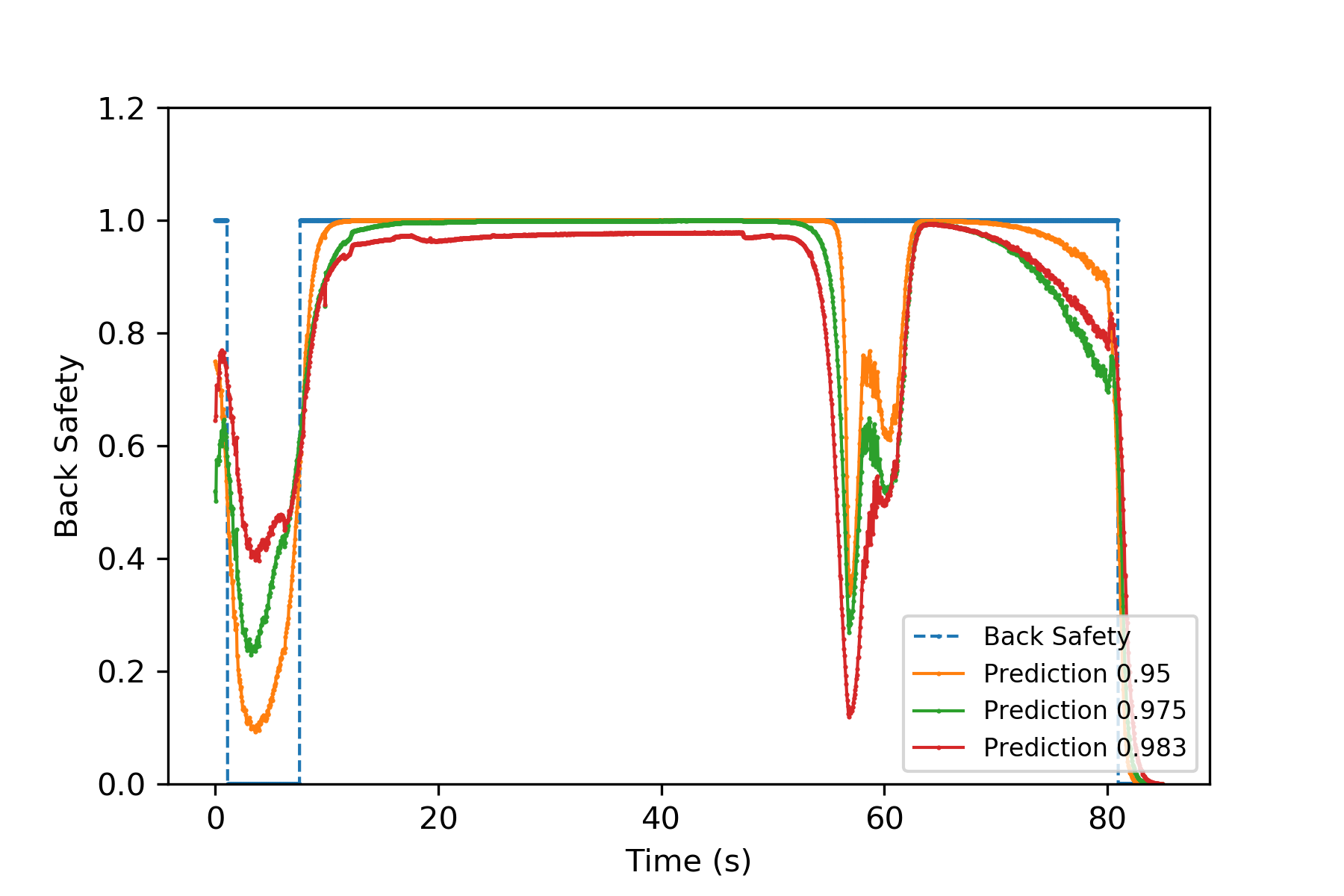}
	}
	\subfloat[Safety distances]
	{
	\includegraphics[width=5.5cm]{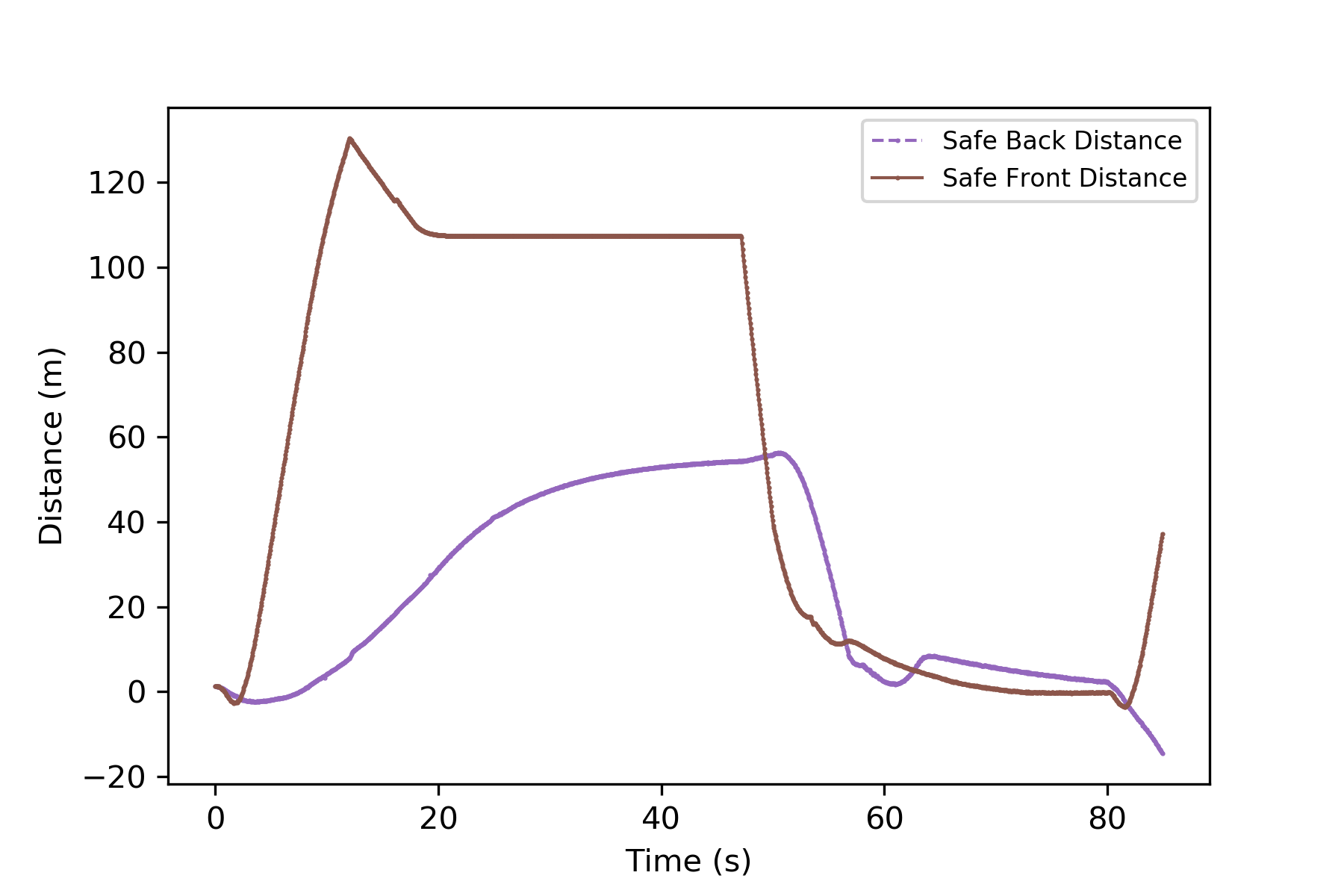}
	}
	
    \subfloat[Front safety predictions]
    {
	\includegraphics[width=5.5cm]{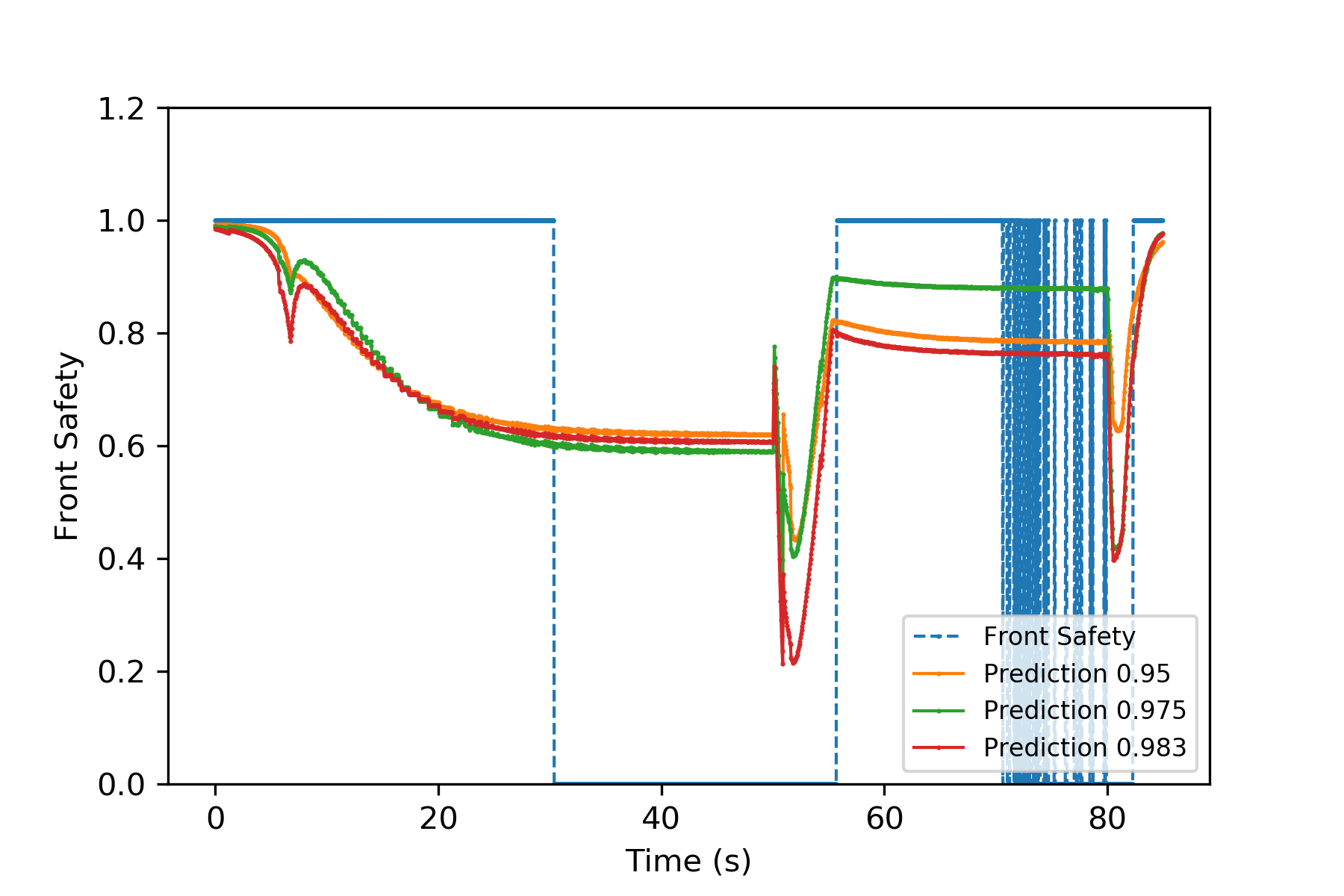}
	}
	\subfloat[Back safe predictions]
	{
	\includegraphics[width=5.5cm]{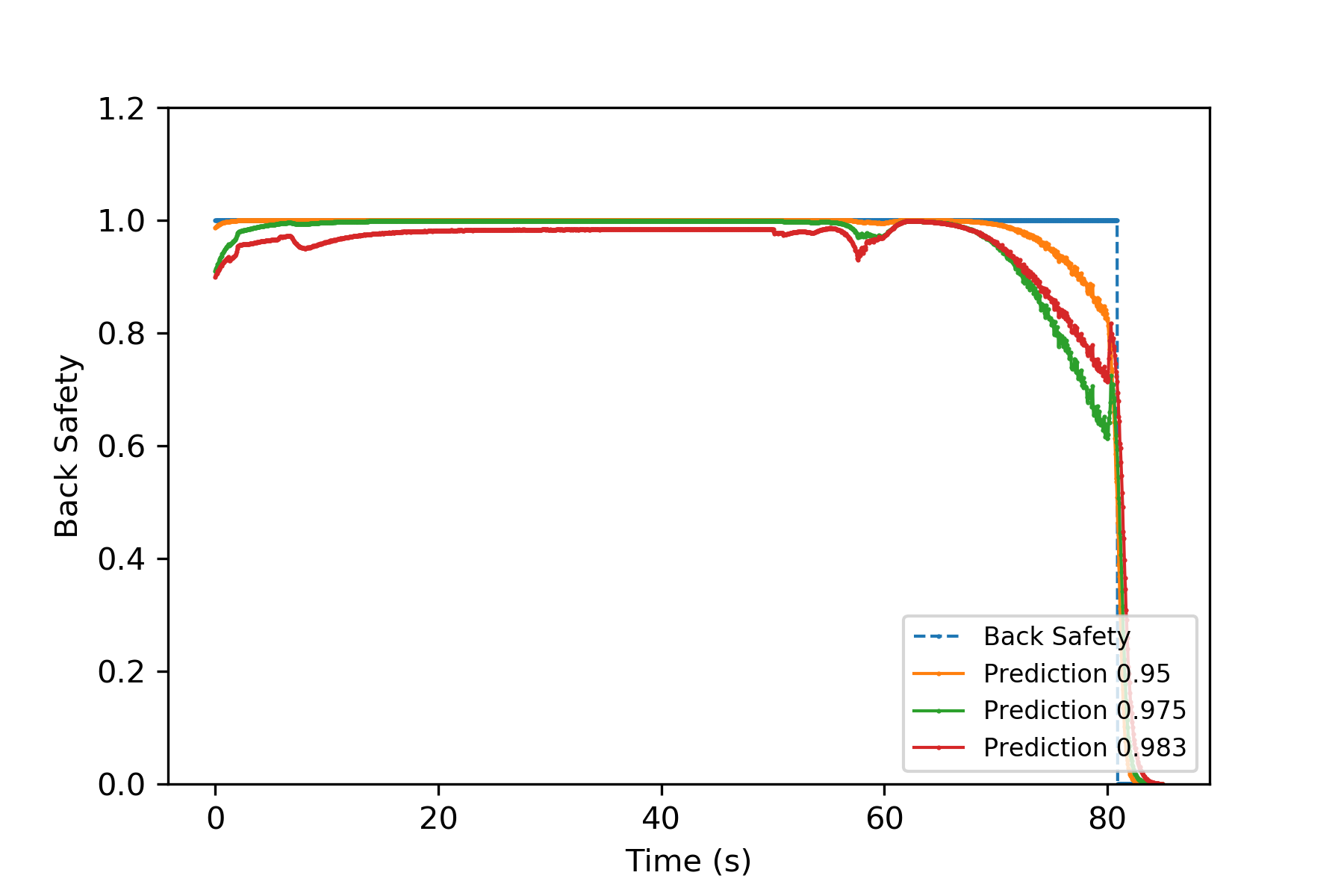}
	}
	\subfloat[Safety distances]
	{
	\includegraphics[width=5.5cm]{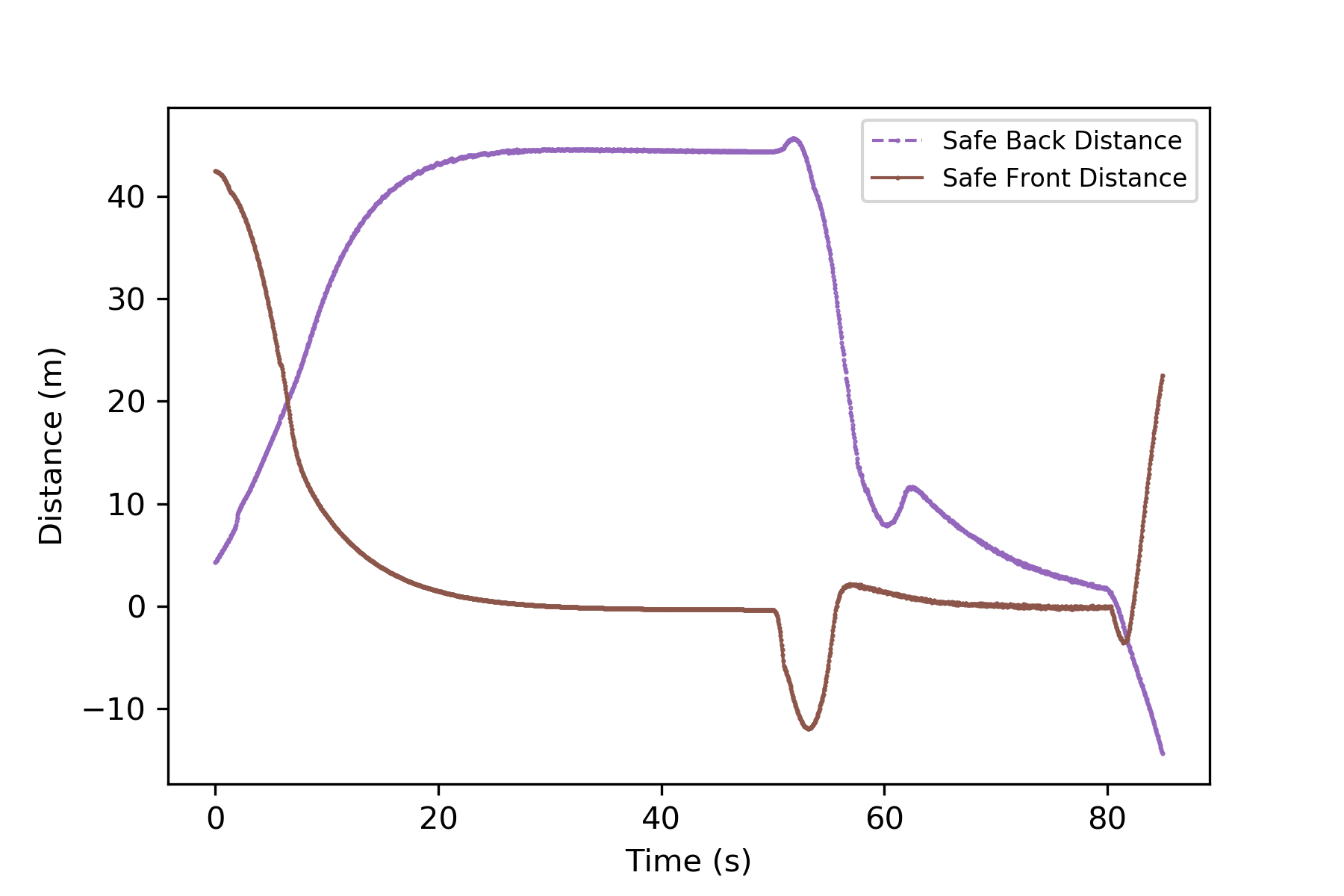}
	}

	\caption{Predicting front and rear safety with different values of $\gamma$ without using the predictions for control.  Top row (a)-(c) is from the emergency stop scenario and the bottom row (d)-(f) is from the follow-and-stop scenario.}
	\label{fig_safety_preds}
\end{figure*}

The safe distances for front and rear are shown in figure \ref{fig_safety_preds}(c) and (f)  for reference.
The safety GVFs predict both front and rear safety effectively and anticipate when the safety could change based on observations of the other agents.
For example, in figure \ref{fig_safety_preds}(b) where the back safety dips at around $t=58$ s, the vehicle is slowing down very quickly while the vehicle behind is not reacting fast enough.
Once the vehicle behind starts to decelerate sufficiently, the predictions jump back up again predicting that the vehicle is safe from a rear-end collision.
The predictive performance is relatively good for all values of $\gamma$.
$\gamma=0.983$ predicts longer term in some cases.
For example in figure \ref{fig_safety_preds}(b), the vehicle starts from a standstill and accelerates with the rear vehicle accelerating quickly behind it resulting in being temporarily unsafe.
Yet the GVF at $\gamma=0.98$ predicting higher values than the other predictors likely because it was predicting longer term into the future.

\begin{figure*}[t]
	\centering
    \subfloat[Safe front distance in emergency stop scenario]
    {
  	\includegraphics[width=5.5cm]{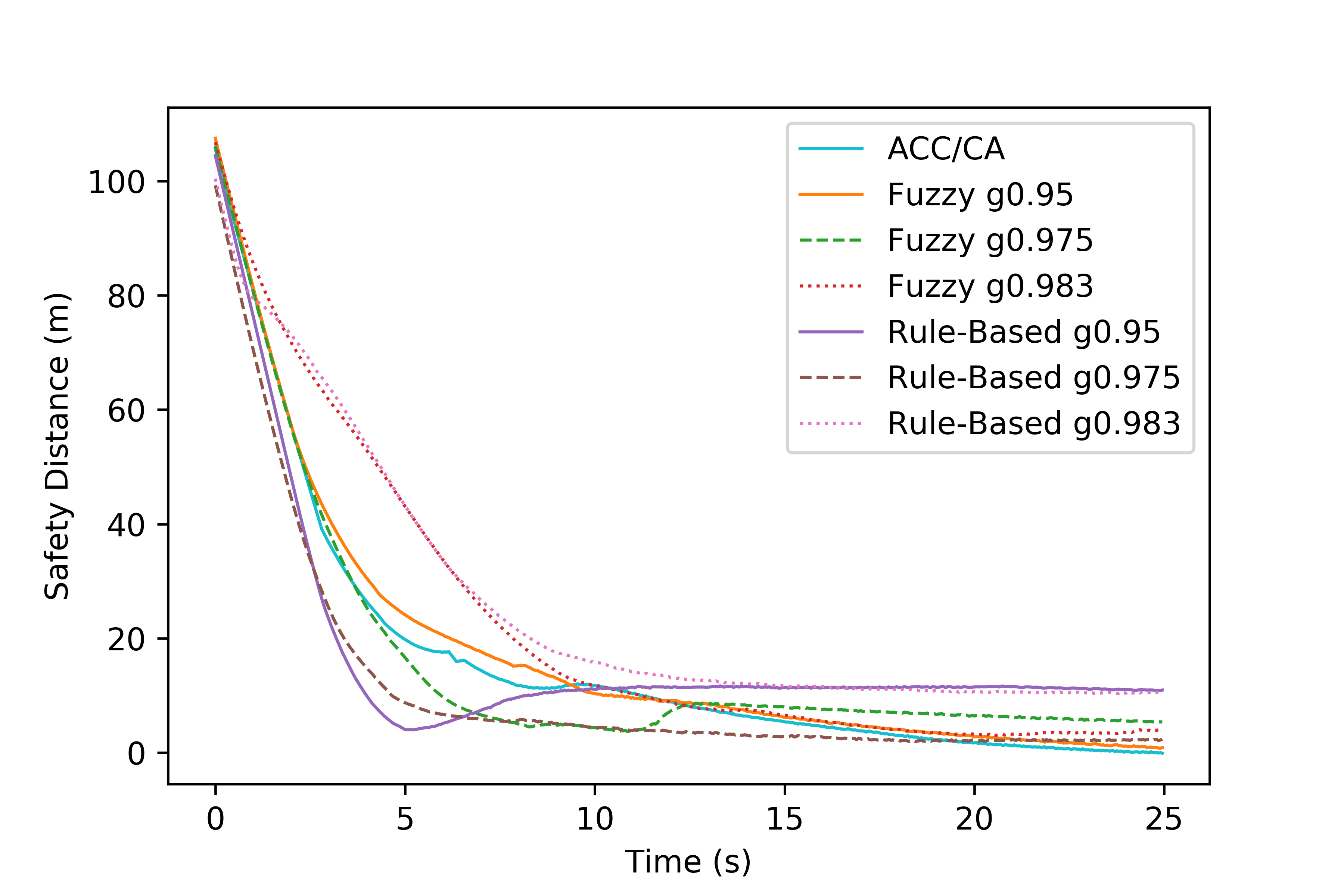}
    }
	\subfloat[Safe front distance in follow-and-stop scenario]
	{
	\includegraphics[width=5.5cm]{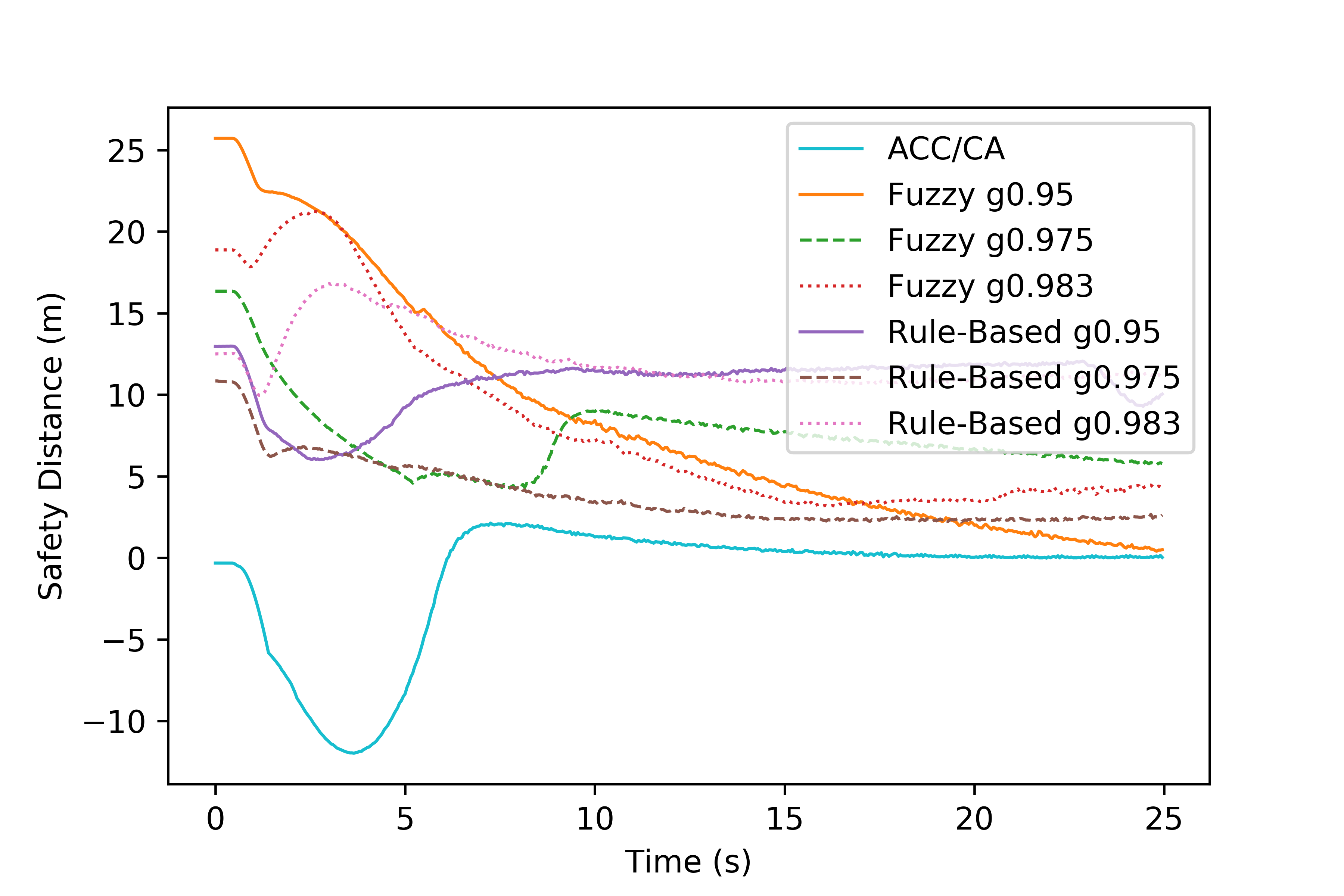}
	}
	\subfloat[Deceleration in follow-and-stop scenario]
	{
	\includegraphics[width=5.5cm]{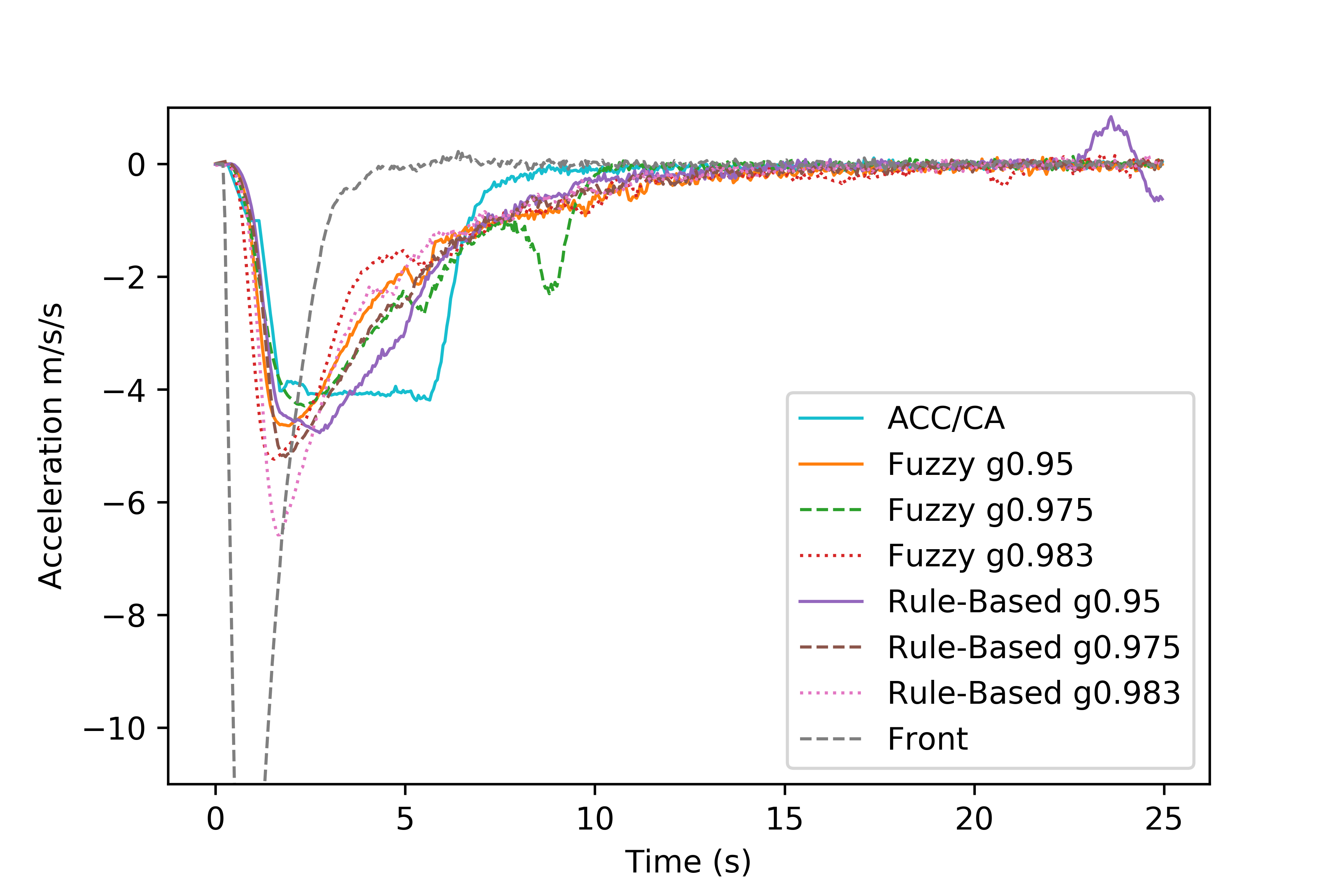}
	}

	\caption{Plots of the deceleration regions of the two scenarios for all values of $\gamma$ and the two GVF-based controllers are given in plots (a)-(c).  (a) The safe front distance is plotted for the emergency stop scenario starting at 100 km/h. (b) Safe front distance is plotted in the follow-and-stop scenario. (c) Deceleration in $\text{m}/\text{s}^2$ the follow-and-stop scenario.}
	\label{fig_control}
\end{figure*}

In figure \ref{fig_control}, the safety distances and acceleration profile is plotted when using the GVF predictions to control the vehicle.
The rule-based and fuzzy-based controllers perform similarly which suggests the predictions learned are useful for the chosen controllers.
The behaviors are not vastly different from the ACC/CA controller which acted as our target baseline.
Our goal was to achieve similar or better performance to ACC/CA to demonstrate that controlling a vehicle with GVF predictions is a viable approach.
From figure \ref{fig_control}(b), the GVF-based controllers appear to optimize the safety rather well and in particular the controllers using $\gamma=0.983$ tend to keep the vehicle safer during deceleration possibly because of being longer-term predictions.
In fact, in most of our experiments, the GVF-based controllers did not cross the safety distance threshold.
It is worth pointing out that sometimes the actual safety did not change despite the predictions changing such as in figure \ref{fig_control}(b); this behavior is due to the ego vehicle stopping quickly while the vehicle behind did not initially apply the brake enough; however, this was eventually rectified to prevent an unsafe situation.
This is also why the longer term prediction $\gamma=0.983$ has a smaller value initially.
It is also noted that while $\gamma=0.983$ weights the future samples more heavily in the prediction, it is harder to learn with TD learning where larger values of $\gamma$ can lead to instability in training.

In terms of computational complexity, the GVF-based approach with the fuzzy controller required only 21 predictions of front safety to determine a suitable next action.
The reason is because the predictions are policy-based rather than action trajectory-based which otherwise would have required a significantly larger search space over all possible action sequences.
We trained an MPC with a non-linear model (not shown here for lack of space) and roughly 3000 predictions were required per time step with an action sequence depth of 5 to achieve similar performance as the proposed approach.
We also note that training a non-linear model was difficult in comparison to the GVF approach but the performance was similar.
We therefore argue that policy-based predictions are a viable way achieve predictive control while keeping computational requirements low.
In addition, the rule-based controller did not require an action search because it relied on the specific choice of target policy to make a policy-based prediction which was used as an error signal in a proportional controller to change the action in the correct direction.

\subsection{Demonstrating in the Real-World}
\begin{figure}[thpb]
\centering
\includegraphics[width=8cm]{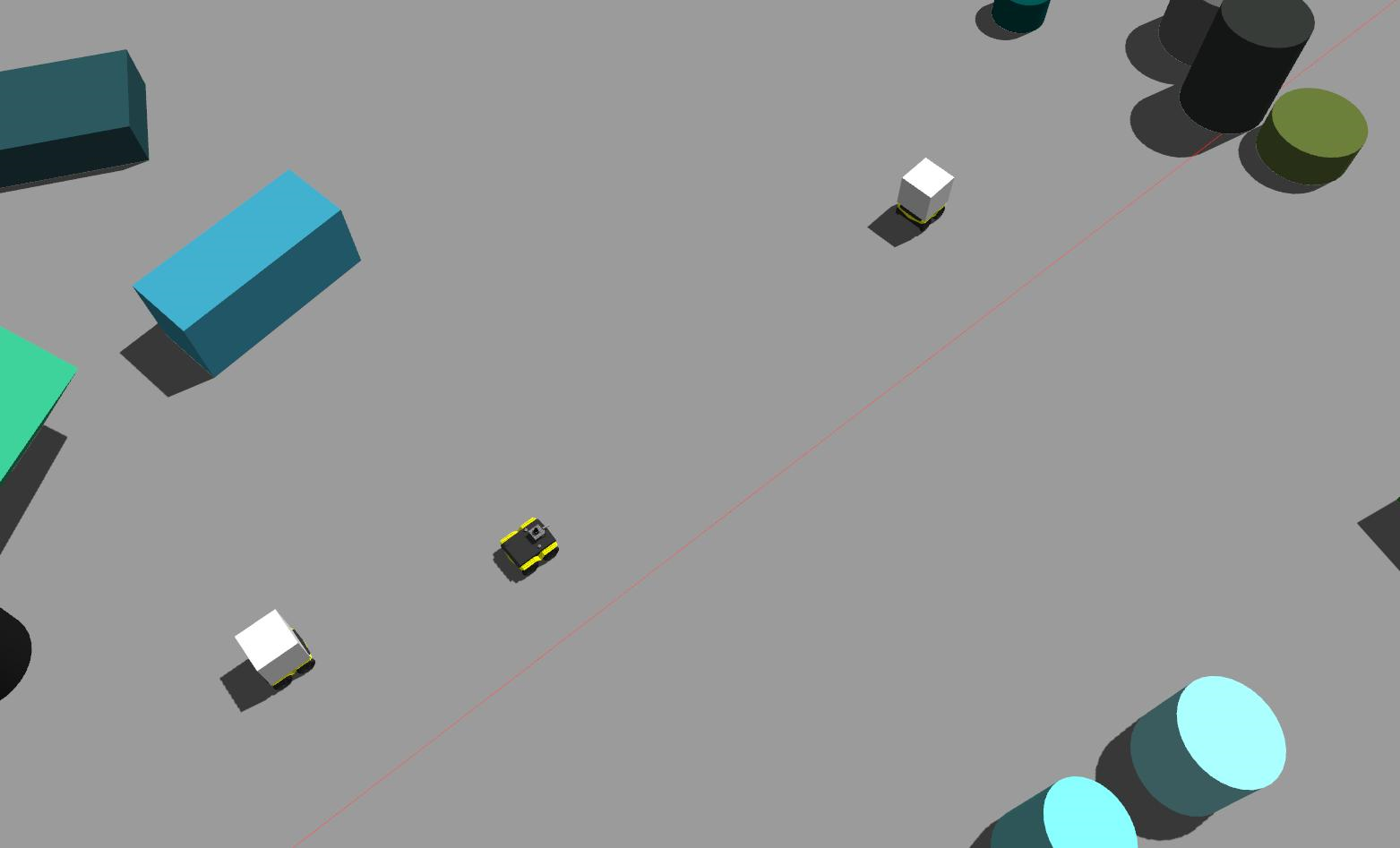}
\caption{Gazebo training environment for the Jackal robot.  This figure shows the ego vehicle and two social vehicles.  The ego vehicle is a simulated Jackal robot with a laser range finder and its sensor readers are used for offline training. We also spawn several additional Jackal robots for social vehicles. The social vehicles have a white cube on them so that the ego vehicle's laser scan can detect them. The simulated Jackals all do a random walk of their linear velocity. Randomly shaped geometric objects are also spawned in the scene to help improve sim-to-real transfer performance.}
\label{fig_gazebo}
\end{figure}

We then trained front safety predictions using a deep convolutional neural network in Gazebo to test on a real-world Clearpath Jackal robot.
In the training environment, randomly shaped geometric objects were generated in the scene to help with sim-to-real transfer to the real-world as shown in figure \ref{fig_gazebo}.

We then tested the predictors on the ClearPath Jackal robot.
The robot has a Hokuyo UTM 30LX laser range finder which produces 1040 distance measurements in a 260$^\circ$ arc in front of the robot.
A 5MP front facing color camera was used to follow blue tape for lateral steering control while the rule-based controller in algorithm \ref{alg_rule_jackal} was used for longitudinal control.
The state vector $s_t$ of the Jackal robot consisted of a history of three LIDAR measurements (each of 1040 points) and current speed $v_t$. 
We used a deep neural network with 6 convolutional layers and 2 fully connected layers to predict safety from the LIDAR measurements.
The safety parameters $\tau=1.5$ seconds and $d_\text{min}=0.4$ meters were used since the robot can stop fairly quickly.
These parameters were chosen because the robot is able to stop very quickly.

The safety predictors were tested on a real robot where we tried several different situations: (a) following a human with varying walking speeds along a pre-defined path, (b) approaching a stationary obstacle, and (c) reacting to a person walking in front of the robot suddenly.
In all cases, the robot was able to stop without collision.
There was some difficulty with interference from strong reflections off some objects in the test area.
This was relatively easy to detect and filter out by replacing measurements less than the minimum range of the sensor with an average range value across the entire measurement.

We also trained front safety predictors in the Webots simulator environment for deploying on an autonomous vehicle in a controlled environment rather than on public roads.
The objects detected in the scene were supplied as input to the neural network which included the distance and speed of the vehicle in front and in the same lane.
We tested the rule-based controller using the predictors and discovered a mismatch between the behavior of the underlying controller that controlled the target speed in simulator and on the vehicle.
The controller on the autonomous vehicle responded slowly which initially created challenges in tuning the proportional controller.
A derivative term was added and this stabilized the control for a comfortable ride.
An emergency brake test was performed where both virtual and real objects were placed in the scene of the vehicle requiring it to stop immediately.
The tests showed that the vehicle was still able to respond safely in situations that required emergency braking.
Finally, we proceeded to test the vehicle in a large circular road where the vehicle had to stop for pedestrians and other vehicles.
The performance was often comfortable and the speed control usually felt human-like.

\section{Conclusions}
In this work, a perception as prediction framework for autonomous driving using GVFs is presented where we focus on predicting safety several seconds into the future.
The predictions learned were both action-oriented and policy-based and it was demonstrated with two different controllers that one could use the predictions to efficiently and safely control a vehicle in the presence of other vehicles and objects.
Based on our experiments, action-oriented safety predictions could be very useful in passive alert systems or in controlling a vehicle.
We believe the safety predictions learned could form the basic building blocks in constructing a more comprehensive control platform in autonomous driving that would rely on learned predictors that better understand how the vehicle's actions and the actions of others can change the environment perceived by the agent's sensors.
We argue that training an agent to understand how actions impact sensory inputs is an important component in controlling an autonomous vehicle.
In fact, this idea is not new.
Model predictive control uses a model that informs the agent how its actions will impact the environment and enables it to find an action trajectory that minimizes a cost function.
The difference is that the GVF predictions are an alternative way to learn and view predictions; one benefit is that they are efficient to use as one prediction is made to describe the final outcome of following a given policy rather than making a sequence of state predictions to evaluate an action.

Finally, we were able to demonstrate that safety predictors learned in simulation can be transferred to a real-world robot and autonomous driving platform to assist in navigating the world safely.
In the future, we plan to extend this work to include safety predictions in multi-lane highways.
If a GVF predictor can predict safety where there are multiple vehicles in multiple lanes, then perhaps the GVF predictor can implicitly understand what a lane is and identify what cues to look for when predicting when a vehicle will change lanes and thereby impact our safety.
In addition, we hope to extend this work to learn predictions off-policy from real-world driving behaviors since the distribution of policies of other agents is hard to model in simulator and better learned from real-world data.

\addtolength{\textheight}{-12cm}





\bibliographystyle{IEEEtran}
\bibliography{IEEEabrv,references}

\end{document}